\documentclass[11pt]{article}
\usepackage{geometry}
\usepackage{coling2020}
\usepackage{times}
\usepackage{bbding}
\usepackage{url}
\usepackage{latexsym}
\usepackage{multirow}
\usepackage{microtype}
\usepackage{graphicx}
\usepackage{amsthm,amsmath,amssymb}
\usepackage{mathrsfs}

\colingfinalcopy 

\title{LMVE at SemEval-2020 Task 4: Commonsense Validation and Explanation using Pretraining Language Model}

\author{ 
  Shilei Liu$^1$, Yu Guo$^1$, Bochao Li$^1$ \and Feiliang Ren$^{12*}$ \\
  $^1$School of Computer Science and Engineering, Northeastern University \\
  $^2$Key Laboratory of Data Analytics and Optimization for Smart \\
  Industry (Northeastern University), Ministry of Education \\
  {\tt \{1901750,1871502,1901725\}@stu.neu.edu.cn} \\ 
  {\tt renfeiliang@ise.neu.edu.cn} \\ 
} 
\date{}

\begin{document}
\maketitle
\begin{abstract}
\blfootnote{\hspace{-0.15cm}$^*$Corresponding author.}

This paper describes our submission\footnote{Our team name on leaderboard is NEUKG, 
and the first three authors contributed equally to this work.} to subtask a and b 
of \emph{SemEval-2020 Task 4}. 
For subtask a, we use a ALBERT based model with improved input form to 
pick out the common sense statement from two statement candidates. 
For subtask b, we use a multiple choice model enhanced by \emph{hint sentence} mechanism 
to select the reason from given options about why a statement is against common sense. 
Besides, we propose a novel transfer learning strategy between subtasks which help improve 
the performance. 
The accuracy scores of our system are 95.6 / 94.9 on official test set\footnote{These scores were obtained during the \emph{Post-Evaluation} phase, and
our scores in the \emph{Evaluation} phase are 90.04 / 93.8.} and rank 7$^{th}$ / 2$^{nd}$ on \emph{Post-Evaluation} leaderboard.

\end{abstract}

\section{Introduction}
\blfootnote{
    %
    %
    %
    %
    \hspace{-0.65cm}  
    This work is licensed under a Creative Commons 
    Attribution 4.0 International Licence.
    Licence details:
    \url{http://creativecommons.org/licenses/by/4.0/}.
    %
    %
}
Common sense verification and explanation is an important and challenging task in
artificial intelligence and natural language processing.
This is a simple task for human beings, because human beings can 
make full use of external knowledge accumulated in their daily lives.
However, common sense verification and reasoning is difficult for machines.
According to Wang~\shortcite{WangLZLG19}, even some state-of-the-art language models such as 
ELMO\cite{peters_deep_2018} and BERT\cite{devlin_bert:_2019}
have very poor performance. So it is crucial to integrate the ability of 
commonsense-aware to natural language understanding model\cite{DBLP:journals/jair/Davis17}.

SemEval-2020 task4\cite{wang-etal-2020-semeval} aims to improve the ability of common sense judgment
for model, and we participated in two subtasks of this task. The dataset of SemEval-2020 task4 named 
ComVE. Each instance in ComVE is composed of 5 sentences $\left \langle s_1, s_2, o_1, o_2, o_3 \right \rangle$. 
$s_1$ and $s_2$ will be used for subtask a, and $s_1$ or 
$s_2$ with $\left \langle o_1, o_2, o_3 \right \rangle$ will be used for subtask b.

Subtask a(also known as \emph{Sen-Making} task)
aims to test a model's ability of commonsense validation. Specifically, 
given two statements $\left \langle s_1, s_2 \right \rangle$ whose lexical and syntactic
are similar, the object of \emph{Sen-Making} model is to determine which statement is
common sense(compared to another one). For example, $s_1$ is
\emph{put the elephant in the refrigerator} and $s_2$ is
\emph{put the turkey in the refrigerator}, a good model needs to judge that 
the latter is more common sense.

Subtask b(also known as \emph{Explanation} task) is a multiple choice task that aims to find the 
key reason why a given statement does
not make sense. For example, given a sentence that violates common sense with three 
options $\left \langle s, o_1, o_2, o_3 \right \rangle$, where 
$s$ is \emph{he put an elephant into the fridge}, 
$o_1$ is \emph{an elephant is much bigger than a fridge}, 
$o_2$ is \emph{elephants are usually gray while fridges are usually white}, and
$o_3$ is \emph{an elephant cannot eat a fridge}, 
the model needs to judge that $o_1$ is the correct option. 

The official baseline of \emph{Sen-Making} use a pretraining language model(PLM) to dynamic encoding the two 
input sentence, and use a simple full connection neural network to calculate the perplexities 
respectively, and choosing the one with lower scores as the correct one. We believe that the 
baseline method treats two sentences independently and ignores the inner relationship between 
the two sentences, so we propose a novel model structure that fully considers the interaction 
between statements.

The official baseline of \emph{Explanation} treats the task as BERT-like multiple choice task\cite{devlin_bert:_2019}. 
We think that the baseline model doesn't make full use of the input data. So we 
design a structure to incorporate another statement that is common sense to existing model.

In addition, we believe that fine-tuning on similar subtask can improve the performance of the 
current subtask because there are many commonalities between the two subtasks, so we propose a 
novel transfer learning mechanism between \emph{Sen-Making} and \emph{Explanation}.

The proposed system named LMVE, it is a neural network model(includes two sub-modules to solve both subtask a and b)
bases on large scale pretraining language model.

Our contributions are as follows:
\begin{itemize}
  \item First, we propose \emph{subtask level transfer learning} that help
        share information between subtasks.
  \item Second, we propose a novel structure to calculate the perplexity of sentence, 
        which takes into account the interaction between sentences in a pair.
  \item Third, we propose the \emph{hint sentence} mechanism that will help
        improve the performance of multiple choice task.(subtask b).
\end{itemize}



\section{System Description}
We consider our model for both \emph{Sen-Making} and \emph{Explanation} as two parts: encoder and decoder.
Encoder is mainly used for getting the contextual representation of input sentence tokens. 
In recent years, some pretraining language models including BERT\cite{devlin_bert:_2019},
RoBERTa\cite{DBLP:journals/corr/abs-1907-11692} and ALBERT\cite{lan_albert_2020} have been proven beneficial
for many natural language processing (NLP) tasks\cite{DBLP:conf/emnlp/RajpurkarZLL16,DBLP:conf/emnlp/BowmanAPM15}.
These
pretrained models have learned general-purpose language
representations on a large amount of unlabeled data, therefore,
adapting these models to the downstream tasks can bring a
good initialization for parameters and avoid training from scratch\cite{xu_improving_2020}.
So we tried some popular PLMs as encoders. 
Decoder consists of several simple linear layers whose number of parameters are 
far less than encoder, and the role of decoder is to fuse the output of encoder and predict the answer.

\subsection{LMVE for \emph{Sen-Making} Task}
\begin{figure}[htbp]
  \centering
  \includegraphics[width=13cm]{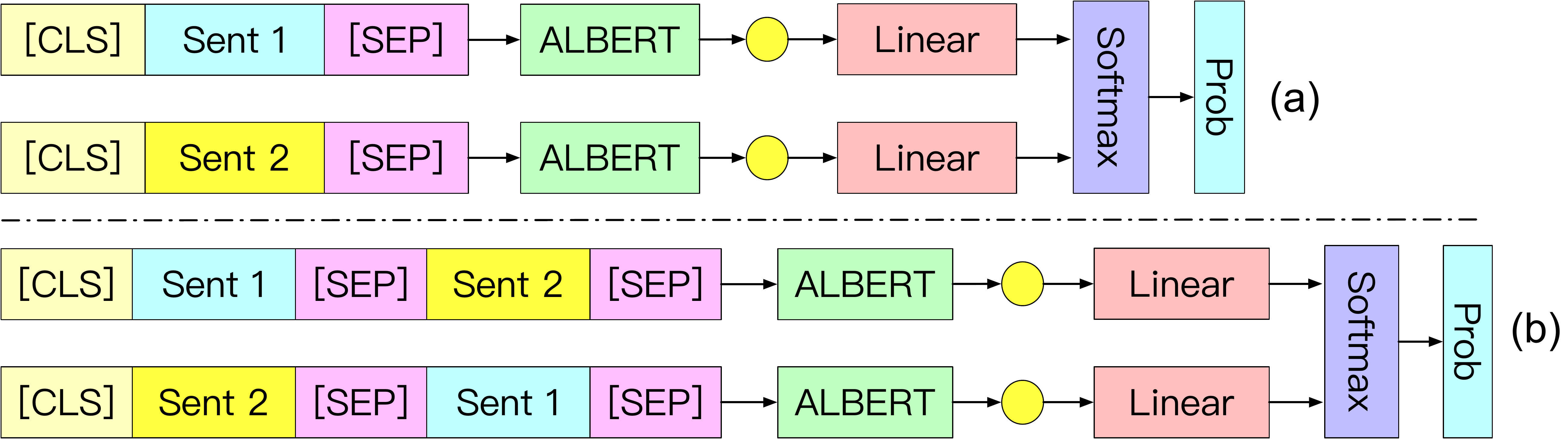}
  \caption{The model architecture for \emph{Sen-Making} task, (a) is official baseline and
    (b) is ours. The yellow point denotes the vector representation of 
    the output sequence(same as below).}
  \label{img:modela}
\end{figure}
Figure \ref{img:modela}(a) is the official baseline model\cite{WangLZLG19}, 
which regards two sentences as independent individuals. But in ComVE, there are 
certain similarities(lexical and grammatical) between the two statements, 
so we think that the interaction between the two sentences is helpful to 
improve the performance of the model.
Figure \ref{img:modela}(b) gives an overview of
our model for \emph{Sen-Making} task which is mainly composed of three
modules including token encoding, feature fusion and answer prediction.

\noindent {\bf Encoding:} Let $\{x^1_0, ... ,x^1_{p}\}$ and
$\{x^2_0, ... ,x^2_{q}\}$ represent one-hot vectors
for the first sentence and the second sentence in an instance, we
first concatenate them and add some special tokens like Figure \ref{img:modela}(b),
then we will get two sequences $\{x_{[CLS]}, x^1_0 , ... , x^1_p, x_{[SEP]},x^2_0 , ... ,x^2_q, x_{[SEP]}\}$ and
$\{x_{[CLS]}, x^2_0 , ... , x^2_q, x_{[SEP]},x^1_0 , ... ,x^1_p, x_{[SEP]}\}$.
The two sequences will be fed into ALBERT respectively.
We use $U^i\in R^{d \times n}$ and $V^i\in R^{d \times n}$ denote outputs
of $i$-th transformer block in ALBERT, where $d$ is the hidden size of model, $n$
is the sequence length, and $i \in \{0,...,L-1\}$.

\noindent {\bf Fusion:} Some pretraining language model(BERT et al.) usually take the first
token (corresponds to {\tt [CLS]}) of the output of last transformer block as the representation
of a sequence, but we use the weighted sum of the representation of first token in all
transformer block outputs as the final representation\footnote{Subsequent experimental results show that the performance using the last 4 layers is the best.}.
The following equations describe the process of fusion:
\begin{align}
  r_i^0      & = U_0^i\qquad r_i^1 = V_0^i                                                                         \\
  \alpha_i & = \frac{\exp(\mathbf{\omega_i})}{\sum_j \exp(\mathbf{\omega_j})} \label{eq:alpha} \\
  x_k        & = \sum_{i=0}^{L-1} \alpha_i r_i^k \label{eq:sum}
\end{align}
where $\mathbf{\omega} \in \mathbb{R}^d$ is a trainable parameter and $i \in [ 0, L-1 ]$.
We can regard $x_k$ as the representation of $k$-th statement($k \in \{0,1\}$).

\noindent {\bf Answer Prediction:} This module maps the output of the fusion
layer to a probability distribution of answer. Given $\mathbf{w}\in \mathbb{R}^d$
and $b \in \mathbb{R}$ as learnable parameters, it calculates the answer
possibility as
\begin{equation}
  p\left(k\right) = softmax \left( \mathbf{w}\boldsymbol{x}_k + b \right) \label{eq:linear}
\end{equation}
We define the training loss as cross entropy loss function:
\begin{equation}
  \mathcal{L}_a=-\frac{1}{N} \sum_{i=1}^N p(y_i)
\end{equation}
where $N$ is the number of samples in the dataset and $y_i \in \{0, 1\}$.

\subsection{Hint Sentence mechanism}
Before formally introducing our model for \emph{Explanation} task, 
let's first introduce the hint sentence mechanism.

In official baseline\cite{WangLZLG19}, the against common sense statement 
will be concatenated with three options respectively and fed into the model. 
We believe that this form of input does not make full use of the data in the ComVE. 
Specifically, for an instance $\left \langle s_1, s_2, o_1, o_2, o_3 \right \rangle$ in ComVE, 
it is assumed that $s_1$ does not conform to common sense, 
then $\left \langle s_2, o_1, o_2, o_3 \right \rangle$ will be used to train 
the baseline model or to predict answer. However, in this process, $s_1$ was abandoned. 
We believe that another common sense statement($s_1$) in statement pair contains 
some useful information, and should be incorporated into our model. 

So we propose hint sentence mechanism: 
A hint sentence is common sense and its lexical and syntactic are similar 
to the given against common sense statement and they differ by only few words. 
In other words, we call the another sentence in the sentence pair a \emph{hint sentence}.

The process of how the \emph{hint sentence} is integrated into the existing model can be 
referred to next section and Figure \ref{img:modelb}. 
The results of ablation experiment(Sec \ref{sec:abla}) show that hint sentence mechanism 
can greatly improve the performance of our model for \emph{Explanation} task.

\subsection{LVME for \emph{Explanation} Task} \label{sec:subtaska}
\begin{figure}[htbp]
  \centering
  \includegraphics[width=14cm]{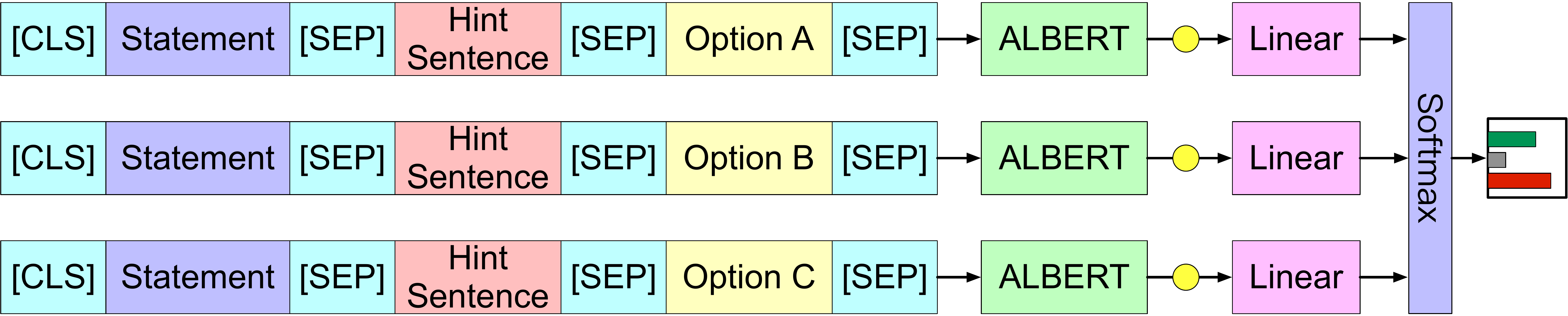}
  \caption{The model architecture for \emph{Explanation} task.}
  \label{img:modelb}
\end{figure}
Figure \ref{img:modelb} gives an overview of our model for \emph{Explanation} task.
it also has three modules.

Let $\{s_0, ... ,s_{p}\}$, $\{h_0, ... ,h_{q}\}$ and $\{o^i_0, ... ,o^i_{r_i}\}$ represent one-hot vectors
for the input statement, hint sentence and $i$-th option in an instance, where $i \in \{0,1,2\}$ and $p$, $q$ and $r$ is the 
length of them, we
first concatenate them and add some special tokens like Figure \ref{img:modelb},
then we will get three sequences $\{x_{[CLS]}, s_0 , ... , s_p, x_{[SEP]},h_0 , ... ,h_q, x_{[SEP]},o^i_0 , ... ,o^i_{r_i}, x_{[SEP]}\}$.
Then the three sequences will be fed into ALBERT respectively.

Similar to last sub-section, each sequence will get a representation vector
after fusion, and then the three representation vector will pass a 
linear layer like Equation \ref{eq:linear} to calculate the
probability distributions of answer.

We define training loss as
\begin{equation}
  \mathcal{L}_b=-\frac{1}{N} \sum_{i=1}^N p(y_i)
\end{equation}
where $N$ is the number of samples in the dataset and $y_i \in \{0,1,2\}$ is
true label.


\subsection{Subtask Level Transfer Learning}\label{sec:tr}
\begin{figure}[htbp]
  \centering
  \includegraphics[width=14cm]{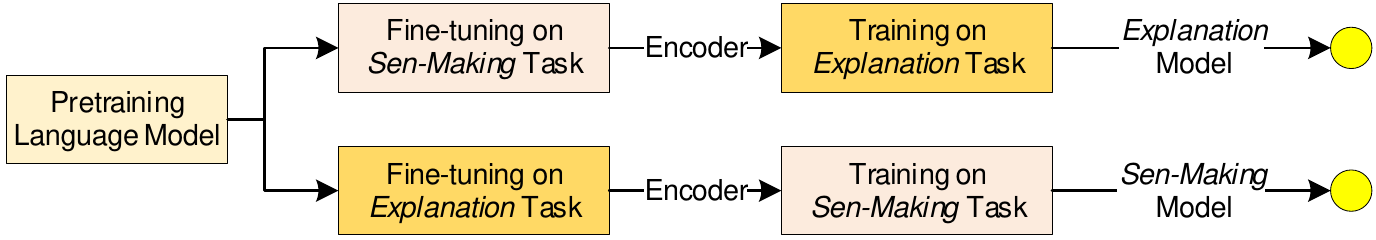}
  \caption{The process of subtask level transfer learning.}
  \label{img:transfer}
\end{figure}
Transfer learning is a research problem in machine learning that focuses on
storing knowledge gained while solving one problem and applying it to a different but related problem.
PLM is a typical example of transfer learning and we call it \emph{task level transfer learning}.

\emph{Sen-Making} task and \emph{Explanation} task are both generalized multiple choice tasks, 
and there is an association between the input data for them, so we believe that in \emph{SemEval-2020 Task 4}, 
fine-tuning on similar subtask can improve the performance
of the current subtask. 

\emph{Subtask level transfer learning} refers to use the encoder after fine-tuning on subtask a(\emph{Sen-Making}) to train
subtask b(\emph{Explanation}) and vice versa.
The process of Subtask level transfer learning are shown in Figure \ref{img:transfer}.

\section{Experiments and Analysis}
\begin{figure}[htbp]
  \centering
  \includegraphics[width=16cm]{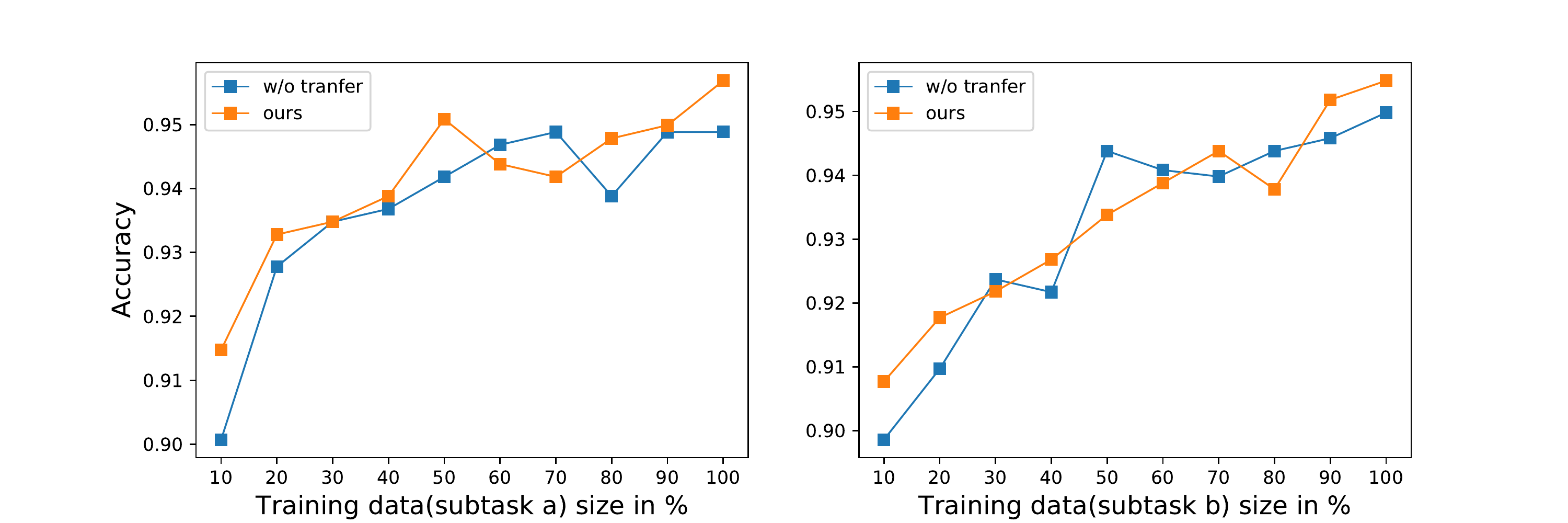}
  \caption{Learning curve on the training dataset.}
  \label{img:plot}
\end{figure}
\subsection{Dataset}
ComVE include 10000 samples in train set and 1000 samples in dev/test set for both \emph{Sen-Making} and 
\emph{Explanation} task. The average length of two statements in the \emph{Sen-Making} task are both 8.26, exactly the same.
The average length of true reasons is 7.63 in \emph{Explanation} task. 

It should be noted that in \emph{SemEval-2020 Task 4}, the test set of \emph{Explanation} task is issued 
only after \emph{Sen-Making} task is completed, so it is impossible to use the test set of \emph{Explanation} task 
to reverse deduce the answer of subtask a test set.

  

\subsection{Baseline}
\label{sec:baseline}
To verify the effectiveness of our model, we used ALBERT to replace the BERT in the 
official baseline, leaving the rest unchanged. We do not perform subtask level transfer
learning (Sec \ref{sec:tr}) on them.
\subsection{Preprocessing}
{\bf Data Augmentation:} To enhance the robustness of
our model, we use Google Sheets\footnote{https://www.google.com/sheets/about} to perform
back translation technology on original texts to get augmented texts.
Specifically, given a training sample
$\left \langle s_1, s_2,o_1, o_2, o_3 \right \rangle$
we first translate the original statements $s_1,s_2$ to
\emph{French} and then translate them back to \emph{English}
(denoted as $\hat{s_1}, \hat{s_2}$).
$\left \langle \hat{s_1}, \hat{s_2}, o_1, o_2, o_3 \right \rangle$
will add to training dataset as a new sample.  the size of the dataset
has doubled after augmentation.


\noindent {\bf Tokenization:} We employ the tokenizer that comes with
the HuggingFace\cite{Wolf2019HuggingFacesTS}
PyTorch implementation of ALBERT. The tokenizer lowercases
the input and applies the SentencePiece encoding\cite{DBLP:conf/acl/Kudo18}
to split input words into most frequent subwords present
in the pre-training corpus. Non-English characters will be removed. 
\subsection{Implementation Details}
We use the \emph{Transformers}\footnote{https://github.com/huggingface/transformers}
toolkit to implemented our model and tune the hyper-parameters according to
validation performance on the development set. The hidden size is equal to
the corresponding PLM. To train our model, we employ the AdamW algorithm\cite{DBLP:conf/iclr/LoshchilovH19}
with the initial learning rate as 2e-5 and the mini-batch size as 48.

We also prepared an ensemble model consisting
of 7 models for \emph{Sen-Making} task and 19 for \emph{Explanation} task with different hyperparameter
settings and random seeds. We used
majority voting strategy to fuse the candidate predictions
of different models together.

\begin{table}[htbp]
  \begin{center}
    \begin{tabular}{llll}
      \hline
      Model                                            & Params    & \emph{Sen-Making} & \emph{Explanation} \\ \hline
      Random                                           & -     & 49.52     & 32.77     \\
      BERT$_{base}$\cite{devlin_bert:_2019}            & 117M     & 88.56     & 85.32     \\
      BERT$_{large}$\cite{devlin_bert:_2019}                                    & 340M     & 86.55     & 90.12     \\
      XLNet\cite{yang_xlnet_2019}                      & 340M      & 90.33     & 91.07     \\
      SpanBERT\cite{DBLP:journals/corr/abs-1907-10529} & 340M      & 89.46     & 90.47     \\
      RoBERTa\cite{DBLP:journals/corr/abs-1907-11692}  & 355M      & 93.56     & 92.37     \\
      ALBERT$_{base}$\cite{lan_albert_2020}            & 12M     & 86.63     & 84.37     \\
      ALBERT$_{large}$\cite{lan_albert_2020}                                 & 18M     & 88.01     & 89.72     \\
      ALBERT$_{xlarge}$\cite{lan_albert_2020}                                & 60M     & 92.03     & 92.45     \\ \hline
      Ours(ALBERT$_{xxlarge}$)                         & 235M     & 95.68     & 95.48     \\
      Our-ensemble                                     & -     & 95.91     & 96.39     \\ \hline
    \end{tabular}
    \caption{Performance with different encoder.}
    \label{tab:encoder}
  \end{center}
\end{table}
\subsection{Main Result}
The result of our model for subtask a and subtask b are
summarized in Table \ref{tab:encoder}. We have tried different
pretraining language model as our encoder, and found that
ALBERT based model achieves the state-of-the-art performance.

Figure \ref{img:plot} shows a learning curve computed
over the provided training data with testing against
the development set, 
and we can see that in the case of \emph{low-resource}
(only use 10\%-20\% training data of target task), the 
performance of introducing subtask level transfer learning 
is significantly higher than original implementation.


\begin{table}[htbp]
  \begin{minipage}{0.48\linewidth}
    \centering
    \begin{tabular}{llll}
      \hline
      \emph{Sen-Making} Model            & Acc & $\Delta_a$ \\ \hline
      Our-single                 & 95.68   & -          \\
      w/o method b$^\dagger$     & 94.88   & 0.80          \\
      w/o data augmentation      & 95.43   & 0.25          \\
      w/o weighted sum fusion    & 95.32   & 0.36          \\
      w/o subtask level transfer & 94.85   & 0.83          \\
      Baseline                   & 93.24   & 2.44          \\ \hline
    \end{tabular}
  \end{minipage}\begin{minipage}{0.48\linewidth}
    \centering
    \begin{tabular}{llll}
      \hline
      \emph{Explanation} Model            & Acc   & $\Delta_b$ \\ \hline
      Our-single                 & 95.48 & -          \\
      w/o hint sentence          & 93.47 & 2.01       \\
      w/o data augmentation      & 95.39 & 0.09       \\
      w/o weighted sum fusion    & 95.11 & 0.37          \\
      w/o subtask level transfer & 94.98 & 0.50          \\
      Baseline                   & 93.12 & 2.36          \\ \hline
    \end{tabular}
  \end{minipage}
  \caption{Ablation study on model components. $\dagger$ means we use the
    model structure as Figure \ref{img:modela}(a) and \emph{Baseline} means 
    the model in Sec~\ref{sec:baseline}.}
  \label{tab:abl}
\end{table}

\subsection{Ablation Study}\label{sec:abla}
To get better insight into our model architecture,
we conduct an ablation study on dev set of ComVE, and the
results are shown in Table \ref{tab:abl}.

From the results we can see that \emph{subtask level transfer learning} has a relatively
large contribution for both subtask a and b, which confirms our hypothesis that
fine-tuning on similar task can improve the performance of the current task. Data
augmentation and weighted sum fusion also have minor contributions due to the
more robust dataset and more robust model.

For subtask a, we can see that compared with baseline method(Figure \ref{img:modela}(a)),
concatenating another sentence as input can have a higher performance. We speculate
that the reason is traditional method treat two statements as independent individuals,
and our method takes into account the inherent connection between the two statements.

For subtask b, we can see from Table \ref{tab:abl} that \emph{hint sentence} makes
a great contribution the overall improvement. We think the reason is a common sense
statement with a similar grammar and syntax does help the model to
determine why the input sentence is against common sense.

\section{Conclusions}
This paper introduces our system for commonsense validation and explanation. For \emph{Sen-Making} task, we use
a novel pretraining language model based architecture to pick out one of the two given statements
that is against common sense. For \emph{Explanation} task, we use a hint sentence mechanism to improve the
performance greatly. In addition, we propose a subtask level transfer learning to share
information between subtasks.

As future work, we plan to integrate the external knowledge 
base(such as ConceptNet\footnote{http://www.conceptnet.io/})  
into commonsense inference.
\section*{Acknowledgements}
We thank the reviewers for their helpful comments. 
This work is supported by the National Natural Science Foundation of China 
(No. 61572120) and the Fundamental Research Funds for the Central Universities 
(No.N181602013).

\bibliographystyle{coling}
\bibliography{semeval2020}

\begin{thebibliography}{}

\bibitem[\protect\citename{Bowman \bgroup et al.\egroup
  }2015]{DBLP:conf/emnlp/BowmanAPM15}
Samuel~R. Bowman, Gabor Angeli, Christopher Potts, and Christopher~D. Manning.
\newblock 2015.
\newblock A large annotated corpus for learning natural language inference.
\newblock In Llu{\'{\i}}s M{\`{a}}rquez, Chris Callison{-}Burch, Jian Su,
  Daniele Pighin, and Yuval Marton, editors, {\em Proceedings of the 2015
  Conference on Empirical Methods in Natural Language Processing, {EMNLP} 2015,
  Lisbon, Portugal, September 17-21, 2015}, pages 632--642. The Association for
  Computational Linguistics.

\bibitem[\protect\citename{Davis}2017]{DBLP:journals/jair/Davis17}
Ernest Davis.
\newblock 2017.
\newblock Logical formalizations of commonsense reasoning: {A} survey.
\newblock {\em J. Artif. Intell. Res.}, 59:651--723.

\bibitem[\protect\citename{Devlin \bgroup et al.\egroup
  }2019]{devlin_bert:_2019}
Jacob Devlin, Ming-Wei Chang, Kenton Lee, and Kristina Toutanova.
\newblock 2019.
\newblock {BERT}: {Pre}-training of {Deep} {Bidirectional} {Transformers} for
  {Language} {Understanding}.
\newblock In {\em Proceedings of the 2019 {Conference} of the {North}
  {American} {Chapter} of the {Association} for {Computational} {Linguistics}:
  {Human} {Language} {Technologies}, {Volume} 1 ({Long} and {Short} {Papers})},
  pages 4171--4186, Minneapolis, Minnesota. Association for Computational
  Linguistics.

\bibitem[\protect\citename{Joshi \bgroup et al.\egroup
  }2019]{DBLP:journals/corr/abs-1907-10529}
Mandar Joshi, Danqi Chen, Yinhan Liu, Daniel~S. Weld, Luke Zettlemoyer, and
  Omer Levy.
\newblock 2019.
\newblock Spanbert: Improving pre-training by representing and predicting
  spans.
\newblock {\em CoRR}, abs/1907.10529.

\bibitem[\protect\citename{Kudo}2018]{DBLP:conf/acl/Kudo18}
Taku Kudo.
\newblock 2018.
\newblock Subword regularization: Improving neural network translation models
  with multiple subword candidates.
\newblock In Iryna Gurevych and Yusuke Miyao, editors, {\em Proceedings of the
  56th Annual Meeting of the Association for Computational Linguistics, {ACL}
  2018, Melbourne, Australia, July 15-20, 2018, Volume 1: Long Papers}, pages
  66--75. Association for Computational Linguistics.

\bibitem[\protect\citename{Lan \bgroup et al.\egroup }2020]{lan_albert_2020}
Zhenzhong Lan, Mingda Chen, Sebastian Goodman, Kevin Gimpel, Piyush Sharma, and
  Radu Soricut.
\newblock 2020.
\newblock \{{ALBERT}\}: {A} {Lite} \{{BERT}\} for {Self}-supervised {Learning}
  of {Language} {Representations}.
\newblock In {\em International {Conference} on {Learning} {Representations}}.

\bibitem[\protect\citename{Liu \bgroup et al.\egroup
  }2019]{DBLP:journals/corr/abs-1907-11692}
Yinhan Liu, Myle Ott, Naman Goyal, Jingfei Du, Mandar Joshi, Danqi Chen, Omer
  Levy, Mike Lewis, Luke Zettlemoyer, and Veselin Stoyanov.
\newblock 2019.
\newblock Roberta: {A} robustly optimized {BERT} pretraining approach.
\newblock {\em CoRR}, abs/1907.11692.

\bibitem[\protect\citename{Loshchilov and
  Hutter}2019]{DBLP:conf/iclr/LoshchilovH19}
Ilya Loshchilov and Frank Hutter.
\newblock 2019.
\newblock Decoupled weight decay regularization.
\newblock In {\em 7th International Conference on Learning Representations,
  {ICLR} 2019, New Orleans, LA, USA, May 6-9, 2019}. OpenReview.net.

\bibitem[\protect\citename{Peters \bgroup et al.\egroup
  }2018]{peters_deep_2018}
Matthew Peters, Mark Neumann, Mohit Iyyer, Matt Gardner, Christopher Clark,
  Kenton Lee, and Luke Zettlemoyer.
\newblock 2018.
\newblock Deep {Contextualized} {Word} {Representations}.
\newblock In {\em Proceedings of the 2018 {Conference} of the {North}
  {American} {Chapter} of the {Association} for {Computational} {Linguistics}:
  {Human} {Language} {Technologies}, {Volume} 1 ({Long} {Papers})}, pages
  2227--2237, New Orleans, Louisiana. Association for Computational
  Linguistics.

\bibitem[\protect\citename{Rajpurkar \bgroup et al.\egroup
  }2016]{DBLP:conf/emnlp/RajpurkarZLL16}
Pranav Rajpurkar, Jian Zhang, Konstantin Lopyrev, and Percy Liang.
\newblock 2016.
\newblock Squad: 100, 000+ questions for machine comprehension of text.
\newblock In Jian Su, Xavier Carreras, and Kevin Duh, editors, {\em Proceedings
  of the 2016 Conference on Empirical Methods in Natural Language Processing,
  {EMNLP} 2016, Austin, Texas, USA, November 1-4, 2016}, pages 2383--2392. The
  Association for Computational Linguistics.

\bibitem[\protect\citename{Wang \bgroup et al.\egroup }2019]{WangLZLG19}
Cunxiang Wang, Shuailong Liang, Yue Zhang, Xiaonan Li, and Tian Gao.
\newblock 2019.
\newblock Does it make sense? and why? {A} pilot study for sense making and
  explanation.
\newblock In Anna Korhonen, David~R. Traum, and Llu{\'{\i}}s M{\`{a}}rquez,
  editors, {\em Proceedings of the 57th Conference of the Association for
  Computational Linguistics, {ACL} 2019, Florence, Italy, July 28- August 2,
  2019, Volume 1: Long Papers}, pages 4020--4026. Association for Computational
  Linguistics.

\bibitem[\protect\citename{Wang \bgroup et al.\egroup
  }2020]{wang-etal-2020-semeval}
Cunxiang Wang, Shuailong Liang, Yili Jin, Yilong Wang, Xiaodan Zhu, and Yue
  Zhang.
\newblock 2020.
\newblock {S}em{E}val-2020 task 4: Commonsense validation and explanation.
\newblock In {\em Proceedings of The 14th International Workshop on Semantic
  Evaluation}. Association for Computational Linguistics.

\bibitem[\protect\citename{Wolf \bgroup et al.\egroup
  }2019]{Wolf2019HuggingFacesTS}
Thomas Wolf, Lysandre Debut, Victor Sanh, Julien Chaumond, Clement Delangue,
  Anthony Moi, Pierric Cistac, Tim Rault, R'emi Louf, Morgan Funtowicz, and
  Jamie Brew.
\newblock 2019.
\newblock Huggingface's transformers: State-of-the-art natural language
  processing.
\newblock {\em ArXiv}, abs/1910.03771.

\bibitem[\protect\citename{Xu \bgroup et al.\egroup }2020]{xu_improving_2020}
Yige Xu, Xipeng Qiu, Ligao Zhou, and Xuanjing Huang.
\newblock 2020.
\newblock Improving {BERT} {Fine}-{Tuning} via {Self}-{Ensemble} and
  {Self}-{Distillation}.
\newblock {\em arXiv:2002.10345 [cs]}, February.
\newblock arXiv: 2002.10345.

\bibitem[\protect\citename{Yang \bgroup et al.\egroup }2019]{yang_xlnet_2019}
Zhilin Yang, Zihang Dai, Yiming Yang, Jaime~G. Carbonell, Ruslan Salakhutdinov,
  and Quoc~V. Le.
\newblock 2019.
\newblock {XLNet}: {Generalized} {Autoregressive} {Pretraining} for {Language}
  {Understanding}.
\newblock In {\em Advances in {Neural} {Information} {Processing} {Systems} 32:
  {Annual} {Conference} on {Neural} {Information} {Processing} {Systems} 2019,
  {NeurIPS} 2019, 8-14 {December} 2019, {Vancouver}, {BC}, {Canada}}, pages
  5754--5764.

\end{thebibliography}

\end{document}